\documentclass{article}
\usepackage{ijcai20}

\usepackage{times}

\usepackage{soul}
\usepackage{url}
\usepackage[hidelinks]{hyperref}
\usepackage[utf8]{inputenc}
\usepackage[small]{caption}
\usepackage{graphicx}
\usepackage{amsmath}
\usepackage{booktabs}
\usepackage[ruled,vlined]{algorithm2e}  
\SetKwInOut{KwInput}{Input} 
\SetKwInOut{KwOutput}{Output} 
\usepackage{amssymb}

\title{Textual Membership Queries}

\author{
Jonathan Zarecki\footnote{Contact Author} \And
Shaul Markovitch
\affiliations
Department of Computer Science, Technion - Israel Institute of Technology
\emails
szarecki@cs.technion.ac.il,
shaulm@cs.technion.ac.il
}

\begin{document}

\maketitle

\begin{abstract}
Human labeling of data can be very time-consuming and expensive, yet, in many cases it is critical for the success of the learning process.
In order to minimize human labeling efforts, we propose a novel active learning solution that does not rely on existing sources of unlabeled data. It uses a small amount of labeled data as the core set for the synthesis of useful \textit{membership queries} (MQs) — unlabeled instances generated by an algorithm for human labeling.
Our solution uses \textit{modification operators}, functions that modify instances to some extent.
We apply the operators on a small set of instances (core set), creating a set of new membership queries.
Using this framework, we look at the instance space as a search space and apply search algorithms in order to generate new examples highly relevant to the learner.
We implement this framework in the textual domain and test it on several text classification tasks and show improved classifier performance as more MQs are labeled and incorporated into the training set.
To the best of our knowledge, this is the first work on membership queries in the textual domain.
\end{abstract}

\section{Introduction}

Machine learning algorithms require sufficient labeled data to produce a high-quality model. However, collecting labeled data can pose a significant challenge in some problem domains and may require human labor.  To reduce the labeling cost, instead of passively accepting and using a labeled training set, the learning algorithm may \emph{actively} request the labeling of instances that are estimated to be useful. 
This approach is called \emph{active learning} (AL). 

One of the first theoretical models proposed for active learning was the \emph{membership queries} (MQs) model \cite{Angluin1988}.
In this setting, the learner may request a label for any unlabeled instance in the instance space, including queries (instances) that the learner generates from scratch.
This approach holds strong theoretical promise as its learning model is more robust than the standard PAC \cite{Valiant1984} learning model in many cases \cite{Bshouty1995,Angluin1987}, since it does not depend on the assumption that the training and testing instances are drawn from the same distribution.

There is one major obstacle, however, when using the MQ model for real-world learning tasks. 
Traditionally, training instances, such as images or natural-language sentences, are mapped into feature vectors.  
MQ-based learning algorithms may generate a query in the form of a feature vector, but a human labeler may not be able to classify it as she can only process an instance (e.g. an image or a sentence). Mapping a feature vector to a recognizable instance may also prove to be very difficult.
Consider for example mapping a vector of word frequencies into a text fragment, or mapping a vector of image features into a recognizable image. 

Additionally, even if a mapping from feature vectors to instances exists, it may map into instances that are outside of the domain of the classifier.
To illustrate this problem, let us look at a flower classification task, where the input is an $N\times N$ black-and-white image of a flower and our goal is to classify it to the correct flower species.
Assume that we use the pixels as binary features, 
thus having $2^{N\times N}$ possible feature vectors.  
In this case, there is an easy mapping from a feature vector to an image, but the vast majority of the resulted images are not of a flower and are therefore not classifiable.
Lang and Baum [\citeyear{Lang1992}] encountered a similar problem in their early attempt to use MQs for handwritten digit recognition.
Given two digits, they combined them into a new image and queried a human oracle for classification. This process, however, often resulted in an unrecognizable digit which the human labeler could not process.
Furthermore, the feature-to-instance mapping is potentially not a valid function, since the mapping induced by the features is not necessarily  1-to-1.
In the textual domain, for example, the bag of words (BOW) feature representation \cite{Harris1954}, is used to represent textual instances. Under this mapping, two different instances may be represented by the same feature vector.
Thus, the feature vector associated with the bag \{bites, dog, man \} may be mapped back to the instances  ``Man bites dog" and ``Dog bites man" with obviously different meanings.

To address these issues, the pool-based \cite{Lewis1994} and stream-based \cite{Cohn1990} approaches for active learning were suggested.
The pool-based approach assumes that a collection (pool) of unlabeled instances is available.  The instances in the pool are evaluated by the learning algorithm using their feature-vector representation, and the instance estimated to be most useful is presented to the labeler.
The stream-based approach assumes the data is presented as a stream of instances, and the learner chooses whether to request the label for the presented instance in a similar fashion.

There are, however, real-world scenarios where using pool-based or stream-based AL can be problematic: 
\begin{enumerate} 
\item A set of unlabeled instances is not available.
\item The learned concept is a very small portion of the instance space, leading to very limited availability of positive examples.  For example,
assume that our task is to find tweets posted by terrorist organizations. As these organizations do not post regularly and not all tweets are found, positive examples can be difficult to find even in large pools of data. 
Another type of such problem domains are tasks where we want to classify a behavior within a specific subgroup, where the total number of examples is small. 
A concrete example of this is offensive language classification in Reddit and its subreddit communities. This behavior is different within every community and we will need specific examples to correctly classify toxic users within each one, it may not be possible to collect enough data in this case.

\item There is a set of unlabeled instances, but the set is too large.
One such example is identifying tweets on a specific subject, as the entire pool of tweets is incredibly large, using pool-based approaches to find relevant examples from the entire Twitter data is impractical, especially when using sophisticated AL methods that are computationally expensive \cite{Houlsby2011BayesianAL,Lindenbaum1999SelectiveSF}.

\end{enumerate}
These problems, associated with pool-based active learning and stream-based active learning, do not affect the MQ approach.
In all of these scenarios, membership queries will be able to generate useful examples while avoiding the issues.

In this work, we present a new general and practical methodology for generating membership queries.
Our work tackles the problems in membership query synthesis and presents a novel algorithm for synthesizing new instances by defining a search space over the set of instances, actively searching for the most informative instances.
Our method is capable of generating near-misses — negative examples that are only slightly different from the positive ones. Such examples have been shown to be helpful for learning \cite{Gurevich2006}.

Our main contributions are as follows:
\begin{enumerate}
\item We present a new, practical way of synthesizing high-quality membership queries by defining a search space over the instance space.
\item We present such a search space over the textual domain, defining its operators and implementing algorithms that utilize them for the generation of new textual membership queries.
\item We present an experimental methodology for testing MQ-related algorithms on real data.
\end{enumerate}

\section{A Framework for Generating Membership Queries \label{sec:textual_membership_queries}}
In this section, we present our framework for generating membership queries. We first formalize the learning setup. 

\subsection{Learning Setup \label{ssec:learning_setup}} 

Let $\Phi$ be a set of instances, called the \emph{instance space}.
Let o: $\Phi\rightarrow\{0,1\}$ be an oracle that can label instances from $\Phi$ as positive (1) or negative (0)\footnote{For simplicity we describe only binary classifications, but our framework can generalize to multi-class problems.}. 
Let $F=f_1,\cdots,f_N$ be a set of feature functions, where $f_i: \Phi \rightarrow D_i$ and $D_i$ is the range of $f_i$.
We denote the space spanned by $F$, $\Psi=D_1 \times\cdots\times D_N$, as the \emph{feature space}.
Let $x\in\Phi$ be an instance. We denote $f_\Psi(x) = \left\langle f_1(x),\cdots,f_N(x) \right\rangle $ as the \emph{feature vector} for $x$. 

It is important to note that we define $o$, the labeling function, to be independent of $F$.  We can replace $F$ by an alternative set of feature functions without affecting the instance labels. Labeling takes place in the instance space while learning takes place in the feature space.
In text classification, for example, a human labeler ($o$)
is only able to classify well-structured sentences and will
find it impossible to label feature-based representations such as Word2Vec \cite{Mikolov2013} embeddings.

\subsection{Membership Query Synthesis in the Instance Space}
Our query synthesis framework assumes that for a given learning task, we are given the following components:
\begin{enumerate}
    \item $C$: a core set of labeled instances.
    \item $K$: the desired number of MQs.
    \item $O$: a set of modification operators on instances, where for each $op \in O$, $op : \Phi \rightarrow \Phi$.  
\end{enumerate}

The quality of the modification operators is crucial to the performance of our algorithm. 
For the algorithm to perform well, the modification operators must be able to create a diverse set of new instances given an input, while keeping their output within the instance space $\Phi$.
Given a set of modification operators $O$, we can define the closure of a given core set $C$ as:
$
cl(C) = \left\{ 
s \mid
\exists c \in C, op_1,\cdots,op_n \in O \lbrack op_n \circ \cdots \circ op_1(c) = s \rbrack 
\right\}
$.
We would like to define the operators such that $cl(C)$ is as large and diverse as possible, such that even with a small core set, $cl(C)$ will be useful as a source for high-utility MQs. 
We would also like a single operator to minimally change the instance, as this will enable the generation of near-miss examples by consecutively applying operators which will slowly move the instance over the classification boundary. 
We will later discuss the implementation of the modification operators for the textual domain.

\subsubsection{Stochastic Query Synthesis \label{s_ssec:rand_gen_alg}}  
A simple way of utilizing the modification operators is to apply them in random to the core set of instances.
The algorithm maintains a set of instances $\Omega$. We initialize $\Omega$ with the core set $C$. At each step, we randomly choose an instance from $\Omega$ and apply a random operator to it. The resulting new instance is added to $\Omega$.
When enough instances have been generated, we return $\Omega \backslash C$ as the MQs.
The pseudo-code for stochastic query synthesis is listed in Algorithm \ref{alg:rand_pool_gen_alg}.

\begin{algorithm}[t]
	\KwInput{$C$ - a core set of initial instances \\ 
    		$O$ - s set of modification operators \\
            $K$ - the required number of new MQs}
	\KwOutput{A set of $K$ membership queries}
	\nl	$\Omega$ = C\;
    \nl	\While{$|\Omega \backslash C| < K$}{
	\nl		new\_inst = Apply(RandSelect($O$),RandSelect($\Omega$)) \;
	\nl		$\Omega$ = $\Omega$ $\cup$ \{new\_inst\} \; 	
    	}
	\nl	return $\Omega \backslash C$ \;
\caption{Stochastic query synthesis}
\label{alg:rand_pool_gen_alg}
\end{algorithm}

\subsubsection{Query Synthesis using Search Algorithms}
The stochastic synthesis algorithm can be improved by treating the instance space as a search space and actively seeking more informative instances to generate.  
We define the instance search space as follows: The state space is the instance space, the actions are the modification operators and the heuristic value is given by an evaluation function.

Possible candidates for such heuristics are existing active-learning functions such as Uncertainty Sampling \cite{Lewis1994}. 
These functions are designed to assign higher values to  feature vectors that are estimated to be more informative. Given such a function, $U_{al}:\Psi\rightarrow\mathbb{R}$, we can compose it  with the feature mapping function $f_\Psi(x): \Phi \rightarrow \Psi$ to get instance evaluation function: $U(x) = U_{al}(f_\Psi(x))$.  

Now we can use any resource-bounded (RB) heuristic search algorithm, such as greedy best first and beam search, to look for informative queries, while enforcing resource-bounds in terms of time and memory.
The search-based MQ-generation algorithm is listed in Algorithm \ref{alg:search_pool_gen_alg}.

\section{Generating Textual Membership Queries}
In this work, we focus on using our approach in the textual domain. At this stage we limit our study to the classification of sentences, but we plan to extend it to larger textual objects.
As discussed in the introduction, generating instances in the textual domain is especially problematic. Synthesized sentences can easily become unreadable when not treated carefully.
Furthermore, common feature representations, such as BOW \cite{Harris1954}, are not surjective: there are feature vectors that do not map to any instance. 
For example, the feature vector \{``man", ``dog", ``cat"\} does not map into any legal sentence.
These representations are also not injective: there is not a one-to-one correspondence between the instance space and the feature space.
As we saw in the introduction, the sentences ``man bites dog" and ``dog bites man" will receive the same BOW vector. 
These limitations prevent us from generating examples in the feature space and mapping them back to the instance space to be tagged.

To apply our methodology to the textual domain, we need to define the instance space and the modification operators (our methodology is independent of the feature space).
We define the instance space to be the set of all syntactically and semantically legal sentences in English.
To define modification operators for textual objects, we first define a \emph{semantic distance function} between terms and then define the \emph{semantic neighborhood} of a term. We will then define modification operators that replace words by other words within their semantic neighborhood. 

\begin{algorithm}[t]  
	\KwInput{$C$ - a core set of initial instances \\ 
    		$O$ - s set of modification operators \\
            $K$ - the required number of new MQs\\
            $U$ - instance evaluation \\
            Search - RB heuristic search algorithm
            }
	\KwOutput{A set of $K$ membership queries}
	\nl	$\Omega$ = C\;
    \nl	\While{$|\Omega \backslash C| < K$}{
	\nl		i = RandSelect($\Omega$)) \;
	\nl		new\_inst = Search(i,O,U) \;
	\nl		$\Omega$ = $\Omega$ $\cup$ \{new\_inst\} \; 	
	}
	\nl	return $\Omega \backslash C$ \;
\caption{Search-based query synthesis}
\label{alg:search_pool_gen_alg}
\end{algorithm}  

\subsection{Computing the Semantic Distance \label{s_ssec:qualitative_kb}} 




For a word to serve as a possible replacement within a specific context, it has to be \textit{functionally similar} \cite{Turney2010} to the original, meaning that the two words \emph{behave} similarly in their context. 
In order for our operators to be able to generate near-misses they should strive to make minimal changes to the instance, thus we will to enforce our operators to suggest words that are also \textit{semantically related}, reducing the change to the sentence from a single operator.

This requirement makes common methods for computing semantic relatedness (such as ESA \cite{Gabrilovich2007} and Word2Vec) unsuitable, as they do not consider functionality.  Thus, "wine" may be considered semantically close to "drinking", despite them having different functionality.  
On the other end, we also considered masked language-models (LM) like BERT \cite{Devlin2019BERTPO} but in many cases found it focused mainly on preserving functional similarity and not on semantic relatedness.
Instead, we chose to use Dependency Word2vec \cite{Levy2014}, a representation with an associated distance function that was specifically trained to exhibit both of these properties.

\paragraph{Modification Operators in the Textual Domain.}
We define a semantic neighborhood of a word $w$ as the set of words that hold related meaning and can be used in similar contexts.
Our modification operators use the semantic neighborhood in order to substitute words in a sentence with other words in their semantic neighborhood, generating new legal sentences. 
Given a semantic distance function,
we define the \emph{k-semantic neighborhood} for a particular word $w$, $N(w,k)$, as the $k$ closest words to $w$.
Using the semantic neighborhood, we can now define the modification operators for a given sentence \emph{s}.
First, all verbs, nouns and adjectives in \textit{s} are marked as \emph{replaceable words}. Then, the k-semantic neighborhood of each replaceable word is calculated.
A candidate operator replaces a \emph{replaceable word} with a member of its k-semantic neighborhood.  The candidate operators are then filtered by Spacy's latest part-of-speech parser to keep only those that retain the syntactic structure of the original sentence.
%
Note that by consecutively applying the modification operators we are able to significantly change the sentence meaning while each operator minimally changes the sentence's overall theme, building a focused set of new sentences more likely to contain near-misses.


%

\begin{figure}
	\centering
  	\includegraphics[width=0.9\linewidth]{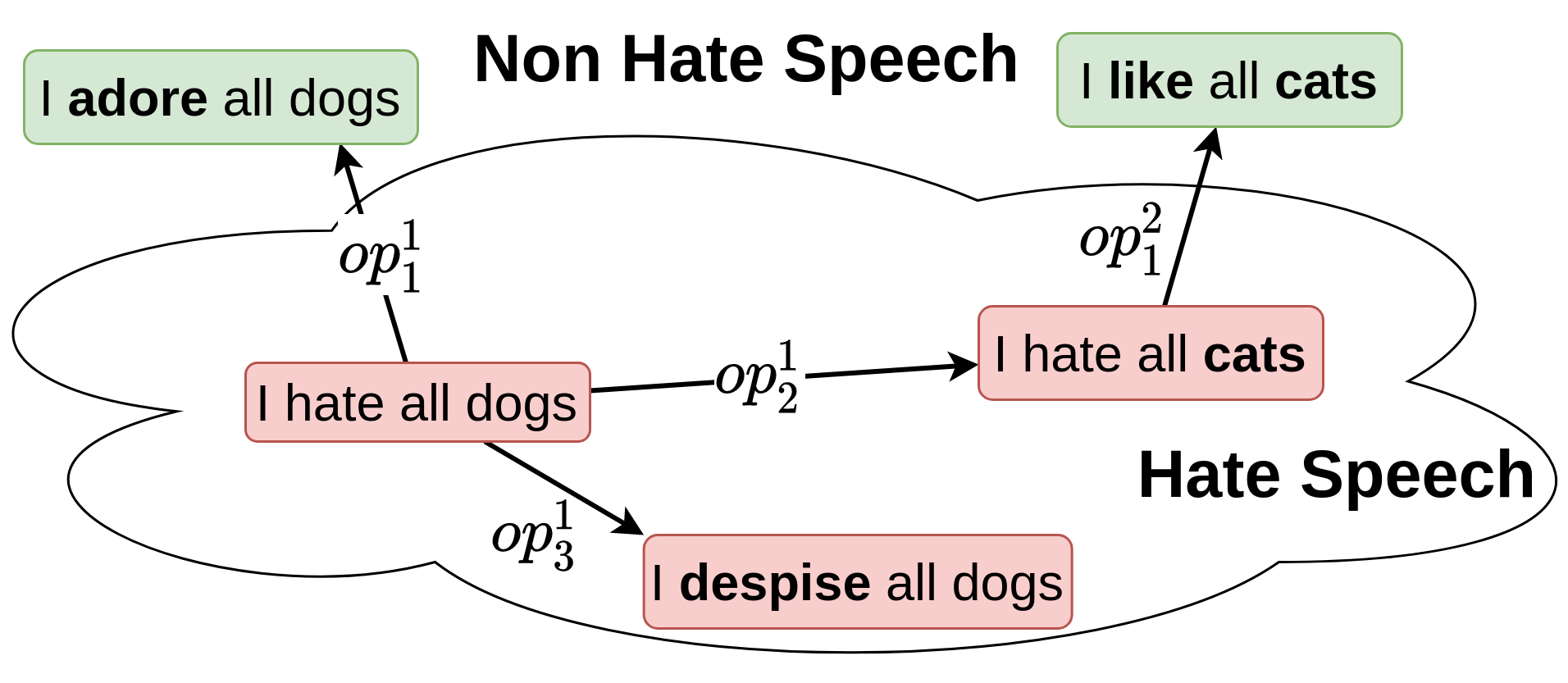}
  	\caption{Modification operators for a sentence in the hate-speech detection task}
  \label{fig:mod_op_text_example}
\end{figure}

\section{Empirical Evaluation}
We analyzed the performance of our framework on 5 publicly available sentence classification datasets. 
The code for all experiments is available here\footnote{www.github.com/jonzarecki/textual-mqs}.

\subsection{Experimental Methodology}


We report results on 5 binary sentence classification datasets: three sentiment analysis datasets, one sentence subjectivity dataset, and one hate-speech detection dataset. 

\noindent  
\textbf{CMR:} Cornell sentiment polarity dataset \cite{Pang2005}.
\textbf{SST:} Stanford sentiment treebank, a sentence sentiment analysis dataset \cite{Socher2013}.
\textbf{KS:} A Kaggle short sentence sentiment analysis dataset.
	\footnote{www.kaggle.com/c/si650winter11} 
\textbf{HS:} Hate speech and offensive language classification dataset \cite{hateoffensive}.
\textbf{SUBJ:} Cornell sentence subjective / objective dataset \cite{Pang2004}.

\subsubsection{Simulating the Human Oracle \label{s_ssec:sim_oracle}}
As in most works on active learning, we require a human expert to label our queries. However, because we generate different MQs in every run and need to label these instances every time, a very significant labeling effort is required. To address this problem, we chose to \emph{simulate} a human labeler \cite{Settles2009}. 
To make the artificial setting as close as possible to a real-world setting, the artificial expert is a classification model trained on the entire labeled dataset. For each dataset, we chose a model close to the state-of-the-art for the task\footnote{For SST, CMR, SUBJ \& KS we used the open-source implementation of Generating Reviews and Discovering Sentiment \cite{RadfordJS17}, which achieved state-of-the-art results for CMR and SUBJ. For KS it achieved 94\% accuracy.
For the hate-speech (HS) dataset we used a BOW-based classifier, which achieved 91\% accuracy.}.
This setup is sound as label information is available only when requesting example labels, as in all AL setups.
The cross-validation accuracy of each artificial expert on its dataset is 86.8\% in CMR, 94.4\% in SUBJ, 86.2\% in SST, 91.0\% in HS, and 94.5\% in KS. The expert achieves close to state-of-the-art performance for each dataset.
One potential problematic aspect of this methodology is that the performance of the artificial expert was tested only on real examples from the datasets, and not on artificially generated instances.  
To account for this, we performed the following experiment.
We used our algorithm to generate a test set of 100 instances for each of the 5 datasets and asked humans to label these instances. We then applied the artificial expert on these test sets.
The accuracy achieved was very close to the one achieved for the real examples (with a maximum of 2\% difference). This confirmed that the artificial expert can operate on our artificial examples.

\subsubsection{Compared Methods}
In the following experiments, we have used 3 variations of our methods:
a) \textbf{Uncertainty sampling hill-climbing MQ synthesis (US-HC-MQ):} The proposed search-based synthesis method, using hill-climbing search and uncertainty sampling-based heuristic function. 
b) \textbf{Uncertainty sampling beam-search MQ synthesis (US-BS-MQ):} Same as (a) but using the beam search algorithm.
\textbf{Stochastic Synthesis (S-MQ):} A degenerated version of our method, described in detail in Section \ref{s_ssec:rand_gen_alg}. 

As no other work attempted to build membership queries in the textual domain prior to this work,
we chose 3 somewhat similar approaches for generation/augmentation of textual instances to compare with:  a) \textbf{Random Original Examples (IDEAL-RAND):} Randomly select a set of unlabeled examples, label them with original labels and insert them into the model. This model has an unfair advantage of using a pool of unlabeled instances not available to the other methods. It is presented to assume an ``lower upper-limit'' role, enabling us to see what would happen if we had access to unlabeled examples;
b) \textbf{RNN Generator (RNN-LM):} A method for generating sentences with an LSTM \cite{Hochreiter1997LongSM} language model. We fine-tuned the models with only the core set of instances. The RNN-LM model was pre-trained on English Wikipedia, and performed conditional generation from the middle of the sentence;
c) \textbf{WordNet-based data augmentation (WNA):} A possible approach to text augmentation, where words are replaced with their respective synonyms from WordNet \cite{Lecun2015}.
All methods used an environment size of 10, and a linear classifier with an average 300-dim GloVe \cite{Pennington2014} word-vectors as the learner.

\subsection{Experiment 1 - Batch Active Learning with Membership Queries}
We measured the performance of our MQ synthesis framework in a batch active learning setup, where at each step, a pool of unlabeled instances is generated, and then \textit{m} (batch size) samples are chosen to be labeled and incorporated into the core set. 
In each step, we use a generation algorithm to return a pool of unlabeled instances with size P. Then a heuristic function U is used to extract the \textit{m} most informative instances in a batch to be labeled by the expert. These labels are then incorporated into the training set and used to train a model. The model's accuracy is then measured against the test set.
We repeat the experiment 20 times with different core sets and report the average results on all available datasets.

\begin{figure}[ht]
	\centering
  	\includegraphics[width=1.0\linewidth]{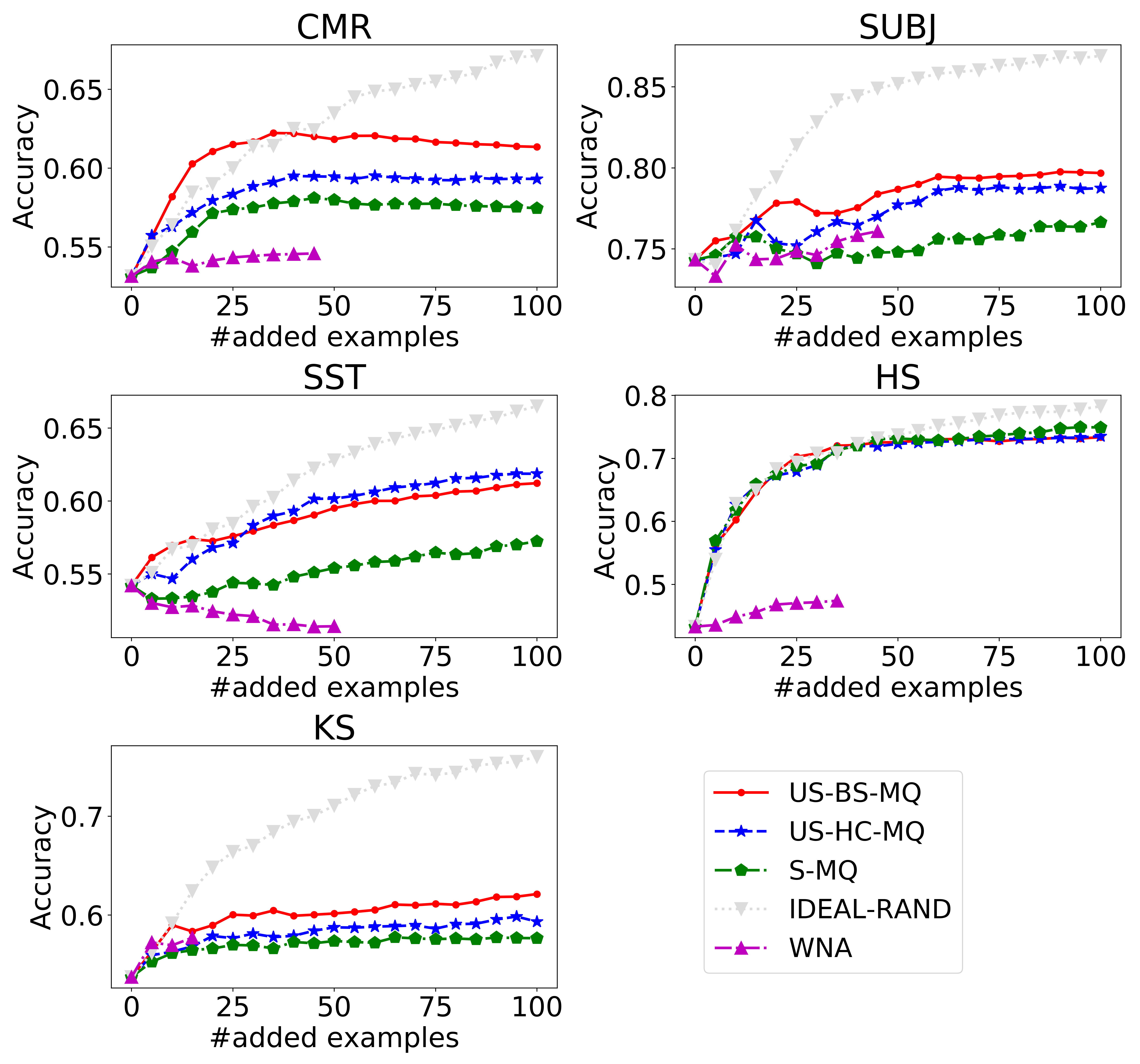}
  	\caption{Comparison of accuracy achieved by the different methods}
  \label{fig:gen_pool_plots}
\end{figure}

Figure \ref{fig:gen_pool_plots} plots the accuracy curves as a function of the number of queries generated by our algorithm or by the competitor methods. We used a core set of 10 sentences, a pool size of 20, an AL batch size of 5, and the uncertainty sampling-based \cite{Lewis1994} heuristic function as U for all experiments. We chose a small core set size to emphasize our algorithm's ability to generate useful sentences even in this challenging setting.

The two search-based approaches (US-HC-MQ and US-BS-MQ) both exhibited excellent performance across the 5 datasets.
The comparison of the search-based approaches to S-MQ showed that, as we expected, more valuable examples are obtained when using the utility function in the generation process.
WNA performed admirably considering that in principle it is using only a small semantic neighborhood and therefore receives only synonyms. However, its lack of diversity resulted in low accuracy on some datasets.
Another significant disadvantage results from WNA using only synonyms is the limited amount of synonyms available from WordNet, making it unable to generate a large pool of instances.
In comparison, our method can theoretically generate as many instances as required.
RNN-LM's results were not shown as many sentences generated using this approach were marked as invalid by human labelers, as this was a requirement for our AL setup we were forced to remove them. Additional information is presented in \ref{ssec:human_eval_sent_gen}.



\subsection{Experiment 2 - The Effect of the Synthesis Algorithm on Label Switches \label{expr:label_switch}} 
In our framework, after applying the modification operators, the resulting instances are able to change their original label and thus cross the classification boundary, becoming ``near-misses'' \cite{Gurevich2006}, examples that originally belonged to a certain class, but our sequence of modification operators caused them to switch their original class without significant changes to the instance.

In this experiment, we tested the effect of our synthesis algorithms on the number of label changes they cause. 
Three algorithms were compared, Uncertainty hill-climbing (US-HC-MQ), Stochastic hill-climbing (S-HC-MQ), and Stochastic synthesis (S-MQ). US-HC-MQ uses a heuristic function to direct its search, S-HC-MQ applies multiple operators randomly, and S-MQ randomly applies only one operator at a time, just as described in Algorithms \ref{alg:rand_pool_gen_alg} and \ref{alg:search_pool_gen_alg}.
We randomly chose a core set of 10 instances for each dataset and used each synthesis algorithm to generate 50 examples.
The score for each algorithm is the portion of examples it generated that crossed the classification boundary.
We repeat the experiment 20 times with different core sets and show the average results on all available datasets. 

\begin{figure}[ht]
	\centering
  	\includegraphics[width=\linewidth]{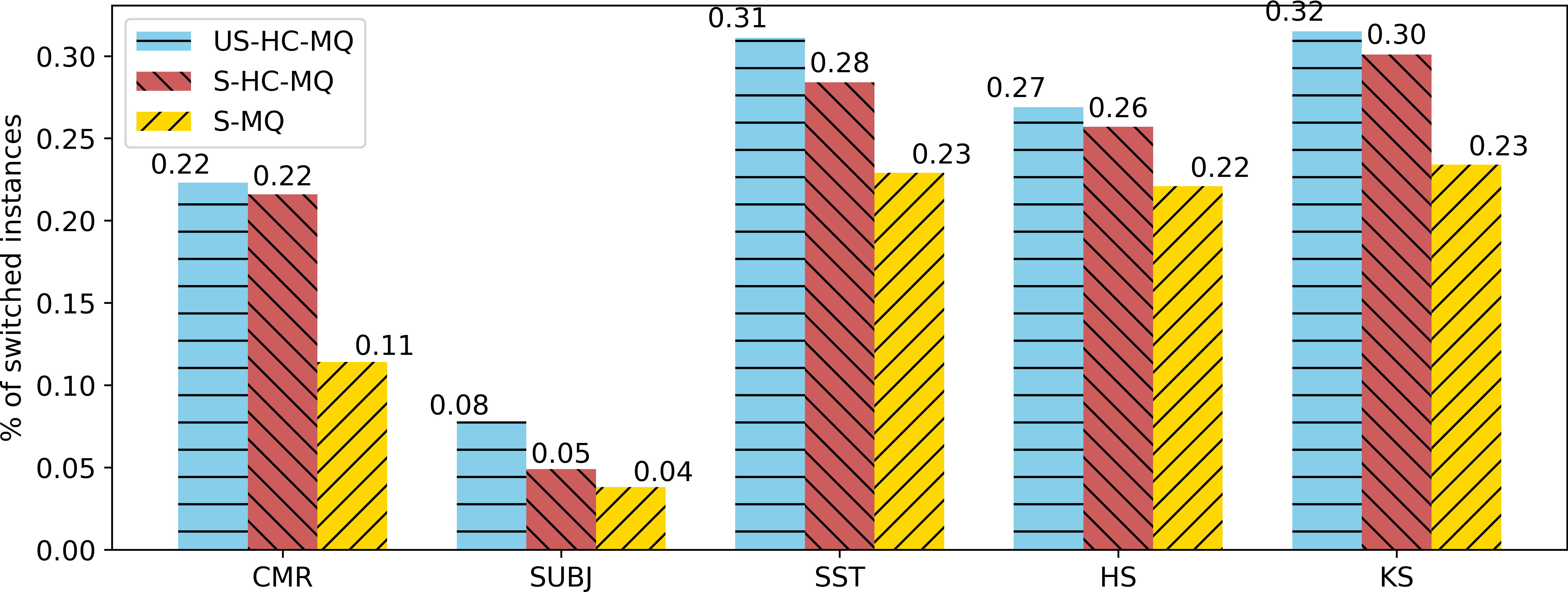}
  	\caption{The effect of the synthesis algorithm on the number of changed labels}
  \label{fig:effect_of_alg_on_label_switch}
\end{figure}

Figure \ref{fig:effect_of_alg_on_label_switch} shows a clear hierarchy, where US-HC-MQ has the most label changes, followed by S-HC-MQ, and then by S-MQ. 
This result reinforces our hypothesis that using multiple operators on a single sentence as well as using heuristic functions during generation results in more diverse sentences as well shows that our framework does not only perform augmentations but significantly changes the meaning of the sentences. 

\subsection{Human Evaluation - Sentence Generation
\label{ssec:human_eval_sent_gen}}
In order to test the quality of our model's generated sentences, we performed a simple readability test using university students as evaluators.
500 sentences generated by our US-BS-MQ, 500 sentences created by the pretrained RNN-LM and 500 original sentences from the data sets used were randomly selected for the test, human evaluators were asked whether each sentence was fully comprehensible to them. 
We performed 3000 of these tests where each sentence was shown twice.
Aggregating the results, 96\% of the original sentences and 95\% of our sentences were labeled as ``comprehensible'', while the RNN-LM's sentences were labeled ``comprehensible'' only 21\% of the time. 
This shows that our artificial sentences are almost indistinguishable from real ones. 
This finding forced us to remove RNN-LM from experiment 1, as it was unable to generate valid instances for labeling, a requirement for an AL setup.

\section{Qualitative Analysis}
Let us look more carefully at the instances generated by our membership queries and the sequence of modification operators that were applied.
Let us look at example sequences: 
\begin{itemize}
\item 
$S_0:$ 
That's why I \textbf{love} Tangled, apart from Moore.
\\ $\downarrow$ \quad $op_1$: love $\rightarrow$ despise \\
$S_1:$ 
That's why I \textbf{despise} Tangled, apart from Moore.

\item 
$S_0:$
I thought Star Wars was really \textbf{boring}.
\\ $\downarrow$ \quad $op_1$: boring $\rightarrow$ fascinating \\
$S_1:$
I thought Star Wars was really \textbf{fascinating}.

\item 
$S_0:$ \quad
Never tell a \textbf{bitch} what u up to. \\
$\downarrow$ \quad $op_1$: bitch $\rightarrow$ slob \quad $op_2$: slob $\rightarrow$ bookworm \\
$S_2:$ \quad
Never tell a \textbf{bookworm} what u up to.

\end{itemize}

These sequences demonstrate the ability of our MQ framework to direct the word switches toward a sentence with a different and sometimes negative meaning, a result seen clearly in Experiment 2.
Now, let us look at an example where our framework did not generate legal sentences.

\begin{itemize}

\item 
$S_0:$
Da Vinci Code sucked, as \textbf{expected}.
\\ $\downarrow$ $op_1$: expected $\rightarrow$ forecasted \quad $op_2$: forecasted $\rightarrow$ preprogrammed \\
$S_2:$
Da Vinci Code sucked, as \textbf{preprogrammed}.
\end{itemize}

In the negative examples, we see that switching a word multiple times can harm the resulting sentence's quality. We see ``expected" being replaced with ``forecasted", a suitable switch, and then with ``preprogrammed" which is not related to the original meaning of the sentence. 
A possible solution is to use multi-sense embeddings \cite{iacobacci2015sensembed} in conjunction with Dependency Word2vec to create different embeddings for each sense of the requested word while preserving the important functional similarity.

\section{Related Work}
In this section we will further discuss three works that are relevant to our method.

\emph{X}-local membership queries were first introduced by Awasthi et al. \cite{Awasthi2012}, defined as a query to any point for which there exists a point in the training sample with a Hamming distance lower than X. Bary [\citeyear{Bary2015}] built on this idea and proved that even a learning model that uses only 1-local MQs (with Hamming distance 1 to a training example) is stronger than the standard PAC learning model.
In addition, Bary also employed a method similar to 1-local queries to gather information for the task of sentiment analysis.

However, neither Bary or Awasthi applied the idea of local membership queries to instances. Rather, they applied this idea to the feature vectors representing those queries. As we discussed in the Learning Setup subsection, it is usually impossible to present an altered feature vector to a human oracle, as was done in these works. Thus the idea of local membership queries has remained mainly theoretical.

Gurevich et al. [\citeyear{Gurevich2006}] presented the idea of modification operators that remain in the instance space. In contrast to the operators applied by Awasthi and Bary, these operators are applied to instances,  easily read by a human expert, and return other instances.
This work used a small seed of only positive instances to model the entire instance space, generating ``near-miss examples" in order to effectively model the vast space of negative instances. 
However, this work was applicable only to the image domain, and its operators obviously could not be applied to textual sentences.
Nor did this work fully explore the options of generation when using the modification operators, such as using search algorithms and heuristics during the MQ generation process.

Several works have discussed the topic of textual data augmentation \cite{Lecun2015,Rosario2017}, where existing examples from the training set are augmented into other very similar instances. However, the augmented instances are not allowed to change their class and so are limited to synonyms which limits the variety of instances they generate.
Indeed, our empirical evaluation showed Zhang \& LeCun's method \cite{Lecun2015} to be less effective than our suggested methods.

\section{Conclusions}
The goal of this work was to show membership queries in a more practical light. We presented a novel \textit{modification operator}-based framework for generating membership queries that are within the instance space and recognizable to the human oracle. Using this framework we presented a local-search-based algorithm for generating MQs that use information from an additional utility function to direct the search to highly valuable instances.
We implemented our approach in the textual domain and demonstrated that our modification operators will result in legal sentences. 
We evaluated our methods on several datasets, finding high accuracy gains when using MQs. Somewhat surprisingly, the approach is sometimes competitive with an approach that utilizes a pool of unlabeled instances not available to our MQ framework.

Our results motivate a few interesting directions for future work that we plan to explore. 
First, our approach can be extended to replacing larger components of text.  In the future, we intend to replace larger grammatical components such as noun phrases and verb phrases.  To do so, however, we first need to develop methods for generating syntactically and semantically well-structured phrases.
Second, our method can be also used for data augmentation as an alternative to methods of oversampling such as SMOTE \cite{Chawla2002} that work in the feature space. By restricting our modification operators to a very tight semantic neighborhood that uses only synonyms, the label of the modified instances is not likely to change, and we can use them for augmentation.
Third, we plan to explore ways to enrich existing pools of unlabeled instances using heuristic-guided local search in a method similar to the one used here, which will result in new high-quality unlabeled examples to choose from.
This direction may enhance the results of existing methods for pool-based AL.

\bibliographystyle{named}
\bibliography{ijcai20}

\end{document}